\documentclass{article}
\usepackage{ijcai16}

\usepackage{times}
\usepackage{amsmath}

\newcounter{examplecounter}
\newenvironment{example}{
\begin{quote}
    \refstepcounter{examplecounter}%

  (\arabic{examplecounter})}
{\end{quote}}

\usepackage{graphicx}

\newcommand{\citep}[1]{\cite{#1}}
\newcommand{\predThF}[1]{{\operatorname{\mathsf{#1}}}}

\title{Robust Natural Language Processing - Combining Reasoning, Cognitive Semantics and Construction Grammar 
for Spatial Language}

\author{Michael Spranger$^1$, Jakob Suchan$^2$ \and Mehul Bhatt$^2$\\
$^1$Sony Computer Science Laboratories Inc., Tokyo, Japan, michael.spranger@gmail.com\\
$^2$University of Bremen, Bremen, Germany, [jsuchan$|$bhatt]@informatik.uni-bremen.de}

\begin{document}

\maketitle

\begin{abstract}
We present a system for generating and understanding of dynamic and static spatial relations 
in robotic interaction setups.  Robots describe an environment of moving blocks using English phrases that include spatial 
relations such as ``across'' and ``in front of''.  We evaluate the system in robot-robot interactions
and show that the system can robustly deal with visual perception errors, 
language omissions and ungrammatical utterances.
\end{abstract}

\section{Introduction}

Spatial language is no doubt important for robots, if they
need to be able to communicate with humans. 
For instance, robots need to be able to understand descriptions
such as the following.

\begin{example}
The block moves across the red region.
\label{e:block-moves-across}
\end{example}

Example \ref{e:block-moves-across} focusses on the path of 
the object \cite{croft2004cognitive}. English speakers also have other means of conceptualizing
movement events. They can, for instance, focus 
on the source of the movement or the goal.

\begin{example}
The block moves from left of you, to right of me.
\label{e:block-moves-from-to}
\end{example}

These examples include various aspects of English language syntax,
semantics and pragmatics \cite{levinson2003space,svorou1994grammars}. 
A complete language processing system for robots 
needs to be able to understand and also generate such utterances. 

Importantly, natural language processing systems need to be robust
against various sources of errors. Humans invariably make
mistakes and robots need to be able to deal with missing or misunderstood words,
grammatical errors etc. At the same time, visual processing of scenes is not
perfect. Objects might be occluded and errors in visual tracking 
might impact tracked paths, and visual recognition of events. Robustness
against visual and language perturbations is crucial. 

\begin{figure}[t]
\begin{center}
\includegraphics[width=.8\columnwidth]{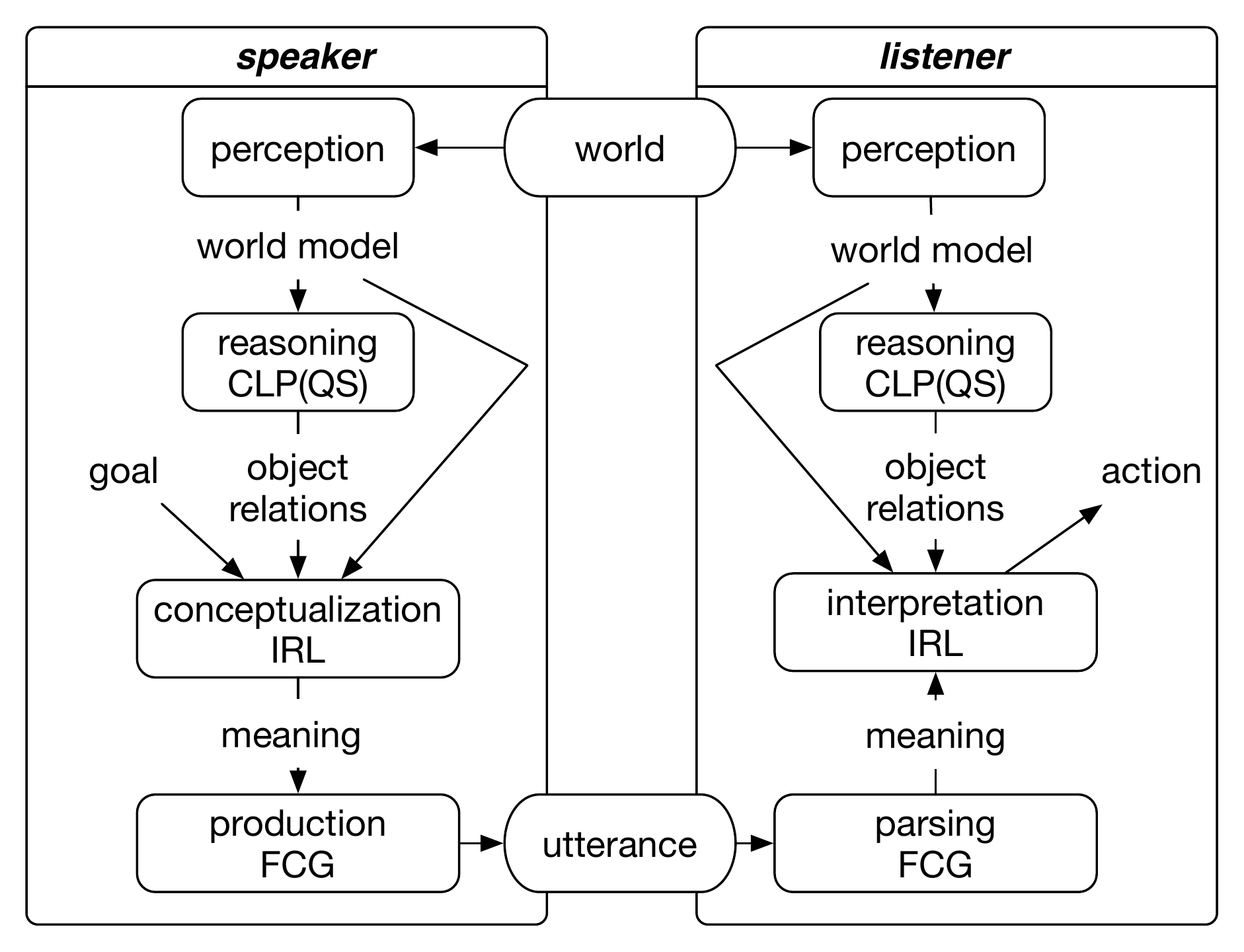}
\end{center}
\caption{Systems involved in processing
of spatial language. Left: processing 
of the speaker to produce an utterance. 
Right: processing of the hearer
for understanding a phrase.}
\label{f:semiotic-cycle}
\end{figure}

In this paper, we present a complete system 
that allows robots to describe and understand descriptions
of spatial scenes involving movement events (see  Figure \ref{f:semiotic-cycle}). 
The system is robust against 
perceptual errors, missing words and grammatical errors.

This paper starts by outlining related work, followed by a
description of main components of the system 
1) a qualitative spatial reasoner that provides qualitative descriptions of dynamic 
scenes,  2) a cognitive semantics system that picks distinctive qualitative aspects 
of a dynamic scene in order to describe events, and 
3) a construction grammar implementation of spatial phrase processing
that allows to produce and parse English sentences. Each of these
components and their integration make the system
robust against various sources of errors. In the final sections we evaluate
the system and show that it is robust against perceptual and language errors.

\begin{figure}[t]
\begin{center}
\includegraphics[width=0.7\columnwidth]{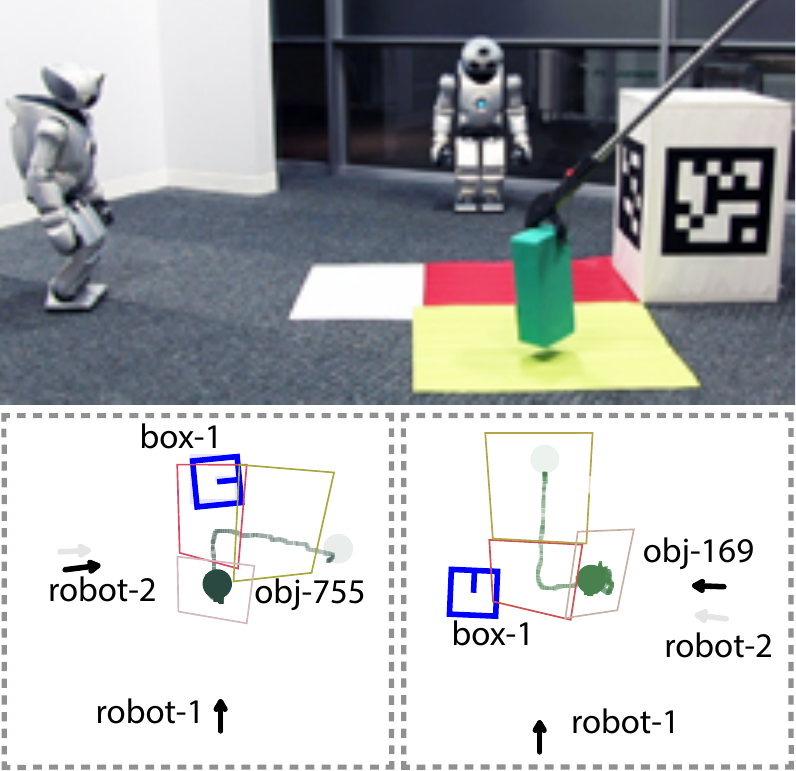}
\end{center}
\caption{Spatial setup. Left: scene model 
extracted by the left robot. Right: 
scene model computed by the right robot. 
Estimated movement of the block (circle) is visualized through opacity. 
The starting point has a lower opacity (alpha). 
Regions are visualized by colored quadrangles. The blue square shows
position, orientation of the box. Arrows are robots. 
{\small\tt robot-1} is the origin of the coordinate system, {\small\tt robot-2} 
position and orientation of the other robot.}
\label{f:setup}
\end{figure}

\section{Related Work}

Earliest systems for spatial language \cite{retz1988various,gapp1995angle,skubic2004spatial} showed
how artificial agents can understand static spatial relations such as ``front'', ``back''. This work has continued. 
We have now various ways of modeling static spatial relations: 
proximity fields for proximal relations \cite{kelleher2006proximity}, prototypes for projective 
and absolute spatial relations \cite{spranger2012deviation}.
Models of static spatial relations are interesting but they 
only cover relations not encoding dynamic qualities.

Recent models of dynamic spatial relations use semantic fields 
\cite{fasola2013using} and probabilistic graphical models \cite{tellex2011approaching} 
for dealing with temporal aspects of spatial relations. In some cases, the work
is on (hand-) modeling spatial relations. Others rely on large task-dependent data sets
in order to learn the representations of spatial relations. 
In general there are fewer approaches using formal methods for
spatial language \cite{spranger2014grounding}.

Formal aspects (e.g., logical, relational-algebraic) and efficient 
reasoning with spatio-temporal information is a vibrant research area within 
knowledge representation and reasoning \cite{ligozat-book}. 
From this perspective, commonsense spatial, temporal, and spatio-temporal relations 
(e.g.,  ``left'',  ``overlap'', ``during'', ``between'', ``split", ``merge'') as 
abstractions for the spatio-linguistic grounding of visual perception and 
embodied action \& interaction have been investigated \citep{Bhatt-Schultz-Freksa:2013-short,Suchan14}. 
Researchers have investigated movement on the basis of an integrated theory 
of space, action, and change \cite{Bhatt:RSAC:2012}, based on theories of time, objects, and position \cite{galton2000qualitative} or defined continuous change using 4-dimensional regions in space-time \cite{muller1998qualitative}.  


One important aspect of robot natural language processing is robustness \cite{bastianelli2014effective}.
Researchers have proposed large coverage, data-driven approaches \cite{chen2011learning},
as well as precision grammar-based approaches for dealing with language problems \cite{cantrell2010robust}.
There are also systems that integrate planning for handling
robustness issues \cite{schiffer2012natural}. More often than not, systems
are evaluated only with respect to natural language errors.
In this paper, we investigate how the integration of formal reasoning methods
with incremental semantic processing and fluid parsing and production
grammars can contribute to robust, grounded  
language processing. 

\section{Grounded Spatial Language Processing}
Two robots interact in an environment 
such as the one shown in Figure \ref{f:setup}. 
For the experiments discussed in this paper, 
we used \emph{Sony humanoid} robots.  The 
vision system of these robots fuses 
information from the robot's camera (30 fps) with
proprioceptive sensors distributed across the body (gyroscope, internal body model
from motor position sensors), in order, to single out and 
track various objects in the environment \cite{spranger2012perception}.

The environment features four types of objects: 
\emph{blocks}, \emph{boxes}, \emph{robots} and \emph{regions}. 
The vision system extracts the objects (as blobs) from the
environment and computes a number of raw, continuous-valued
features such as \emph{x}, \emph{y} position, \emph{width}, and 
\emph{height} and colour values (YCbCr). Objects are tracked 
over time and assigned unique identifiers as long as there is
spatio-temporal continuity. For instance, the green block has been given
the arbitrary id {\small\tt obj-755} by the left robot. 

\subsection{Reasoning about Space and Motion}

The robots generate qualitative representations of the spatio-temporal dynamics in the scene as perceived by their vision system. Towards this, we use a general theory of space and motion implemented based on CLP(QS) \cite{Bhatt-clpqs-cosit11} - a \emph{declarative spatial reasoning framework}, which implements declarative spatial relations in constraint logic programming within the PROLOG programming environment. We use the framework for defining events grounded in the visual observations of the robots, using qualitative spatial and temporal relations between objects in the scene, i.e. \emph{topology, orientation, and movement}.

\begin{figure}[t]
  \centering
  \includegraphics[width=0.9\columnwidth]{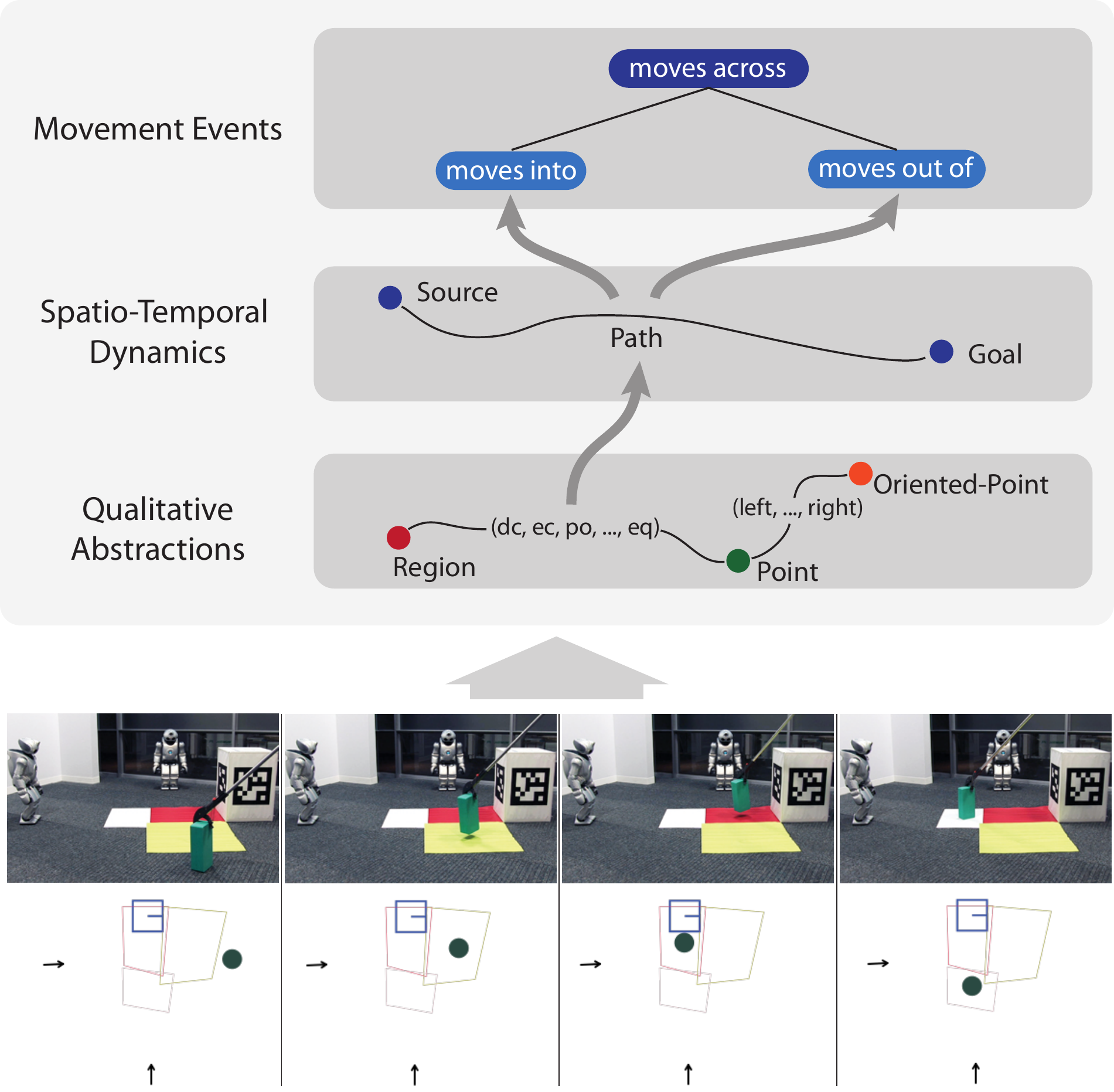}
      \caption{Dynamic scene as observed by the robot to the left, and representation of spatio-temporal dynamics in qualitative abstractions of \emph{space}, \emph{time}, and \emph{motion}. } 
  \label{fig:dynamic_data}
\end{figure}

In order to reason about the perceived dynamics of scenes (for example the scene in Figure \ref{f:setup}), we generate sequences of movement events based on the perceptual data of the robots, as depicted in Figure \ref{fig:dynamic_data}. 
Towards this, objects are represented using \textbf{qualitative abstractions} of spatial 
properties, e.g. position, orientation, extend in space, using 
\emph{primitives} such as \emph{regions}, \emph{points}, \emph{oriented points}, 
\emph{line segments}. Perceived \textbf{spatio-temporal dynamics}, i.e. the movement of the block is represented by the \emph{source} and the \emph{goal} of the movement, and the \emph{path}, on which the object moves from the \emph{source} to the \emph{goal}.  For describing the movement and involved \textbf{movement events}, we use spatio-temporal relations, e.g. for representing the \emph{source} and \emph{goal} locations of the movement with respect to the observing robots or the characteristics of the \emph{path}.

The spatial configuration of objects in the scene is represented 
using $n$-ary \emph{spatial relations} $\mathcal{R}\ = \{r_1, r_2, ... ,r_n\}$, in particular, 
we use topological relations of the RCC8 fragment of the RCC calculus \citep{randell92RCC}, 
{\footnotesize $\mathcal{R}_{\mathsf{top}} \equiv {\{}\mathsf{dc}$, $\mathsf{ec}$, $\mathsf{po}$, $\mathsf{eq}$, $\mathsf{tpp}$, $\mathsf{ntpp}$, $\mathsf{tpp^{-1}}$, $\mathsf{ntpp^{-1}}\}$} and orientation relations of the $\mathcal{LR}$ calculus \cite{Scivos2005} {\footnotesize $\mathcal{R}_{\mathsf{orient}} \equiv {\{}\mathsf{l}$, $\mathsf{r}$, $\mathsf{i}$, $\mathsf{s}$, $\mathsf{e}$, $\mathsf{f}$, $\mathsf{b}$\}}. 
Predicates $\predThF{holds-at}(\phi, r, t)$ and $\predThF{holds-in}(\phi, r, \delta)$  are used to denote that the fluent $\phi$ has the value $r$ at time point $t$, resp. in the time interval $\delta$. 
Movement events are used to describe spatio-temporal dynamics of the perceived scene, i.e. how the spatial configuration of objects changes during the movement of the block. 
We use the predicate $\predThF{occurs-in}(\theta, \delta)$  to denote that an \emph{event} $\theta$ occurred in a time \emph{interval} $\delta$.

In particular, movement events are defined by spatio-temporal relations holding between the involved objects and changes within these relations, happening as a part of the event, using the relations of Allen's interval algebra \cite{Allen1983} 
{\footnotesize ${\{}\mathsf{before},$ $\mathsf{after},$ $\mathsf{during},$ $\mathsf{contains},$ $\mathsf{starts},$ $\mathsf{started\_by},$ $\mathsf{finishes},$ $\mathsf{finished\_by},$ $\mathsf{overlaps},$ $\mathsf{overlapped\_by},$ $\mathsf{meets},$ $\mathsf{met\_by},$  $\mathsf{equal}{\}}$} 
for representing temporal aspects of the event. E.g. the event $\predThF{moves\_into}$, representing that a block moves into a region is defined as follows.

\noindent\begin{minipage}{\columnwidth} 
{
\small
\begin{align}\raisetag{6pt}
\begin{split}
&\predThF{occurs-in}(\mathsf{moves\_into}(o_1, o_2), \delta) \supset\\
&\quad \predThF{holds-at}(\phi_{top}(position(o_1), region(o_2)), \mathsf{outside}, t_{1})~\wedge\\
&\quad \predThF{holds-at}(\phi_{top}(position(o_1), region(0_2)), \mathsf{inside}, t_2)~\wedge\\
&\quad \predThF{starts}(t_{1}, \delta) \wedge \predThF{finishes}( t_2, \delta)~\wedge \predThF{meets}( t_{1}, t_2).
\end{split}
\end{align}
}
\end{minipage}

Accordingly, movement events describing a range of perceivable spatial changes can be defined, e.g. moves to, moves across, etc.
Complex interactions can be described by combining multiple movement events.  

To describe the dynamics observed by one of the robots we generate a temporally-ordered sequence of movement events. 
E.g. the following \textbf{Movement Sequence} ($\Psi$) describes the movement in a scene (Figure \ref{f:setup}), 
as observed by the robot to the left. 

\noindent\begin{minipage}{\columnwidth} 
{\small
\begin{align}
\begin{split}
\Psi ~\equiv ~&\predThF{occurs-in}(\mathsf{moves\_into}(obj\operatorname{-}755,~reg\operatorname{-}36), \delta_1)~\wedge\\
&\predThF{occurs-in}(\mathsf{moves\_out\_of}(obj\operatorname{-}755,~reg\operatorname{-}36), \delta_2)~\wedge\\
&\predThF{occurs-in}(\mathsf{moves\_across}(obj\operatorname{-}755,~reg\operatorname{-}36), \delta_3)~\wedge\\
&\predThF{occurs-in}(\mathsf{moves\_into}(obj\operatorname{-}755,~reg\operatorname{-}37), \delta_4)~\wedge\\
&\predThF{occurs-in}(\mathsf{moves\_out\_of}(obj\operatorname{-}755,~reg\operatorname{-}37), \delta_5)~\wedge\\
&\predThF{occurs-in}(\mathsf{moves\_across}(obj\operatorname{-}755,~reg\operatorname{-}37), \delta_6)~\wedge\\
&\predThF{occurs-in}(\mathsf{moves\_into}(obj\operatorname{-}755,~reg\operatorname{-}38), \delta_7).
\end{split}
\end{align}}
\end{minipage}

To reason about the possibility of a movement event to happen at a certain time point, we introduce predicates to describe in which spatial situations an event might happen, i.e. we use the predicate $\predThF{poss-at}(\theta, t)$, to describe the spatial preconditions of an event.

\noindent\begin{minipage}{\columnwidth} 
{\small
\begin{align}
\begin{split}
&\predThF{poss-at}(\mathsf{moves\_into}(o_1, o_2), t) \supset\\
&\quad \predThF{holds-at}(\phi_{top}(position(o_1), region(o_2)), \mathsf{outside}, t).
\end{split}
\end{align}}
\end{minipage}

Further, we use the predicate $\predThF{causes}(\theta ,\phi, r)$ to describe how an event changes the spatial configuration in the scene.

\noindent\begin{minipage}{\columnwidth} 
{\small
\begin{align}
\begin{split}
&\predThF{causes}(\mathsf{moves\_into}(o_1, o_2),\\
& \quad \phi_{top}(position(o_1), region(o_2)), \mathsf{inside}).
\end{split}
\end{align}}
\end{minipage}

These predicates are used to reason about whether an event is a possible subsequent event given
observed events.

\noindent \textbf{Mechanisms for Robustness} 
The reasoning system abstracts from the numerical values of the visual data stream, thereby generalizing observations.
Consequently, small perceptual errors have less or no effect on computed movement events. Similarly, missing observations have little 
effect on the extracted movement sequence, as long as there is at least one observation for each 
qualitative state. For example, for $\predThF{moves\_into}$ only one observation outside the region 
and one observation inside the region is needed. Lastly, reasoning about the 
possibility of movement events increases the chances of agreement between two robots. 
E.g. if a robot observes a  $\predThF{moves\_into}$ event in
a particular region, the robot can reason, that the next possible event could be a $\predThF{moves\_out\_of}$ 
event from that region. The possibility of a $\predThF{moves\_out\_of}$ event together with the observed $\predThF{moves\_into}$ 
leads to the possibility of a $\predThF{moves\_across}$ event. If now he hears from the other robot that there was a
$\predThF{moves\_across}$ event - he can conclude that this is a possible description (taking into account that there might have
been perception errors).

\begin{figure}[t]
\begin{center}
\includegraphics[width=1\columnwidth]{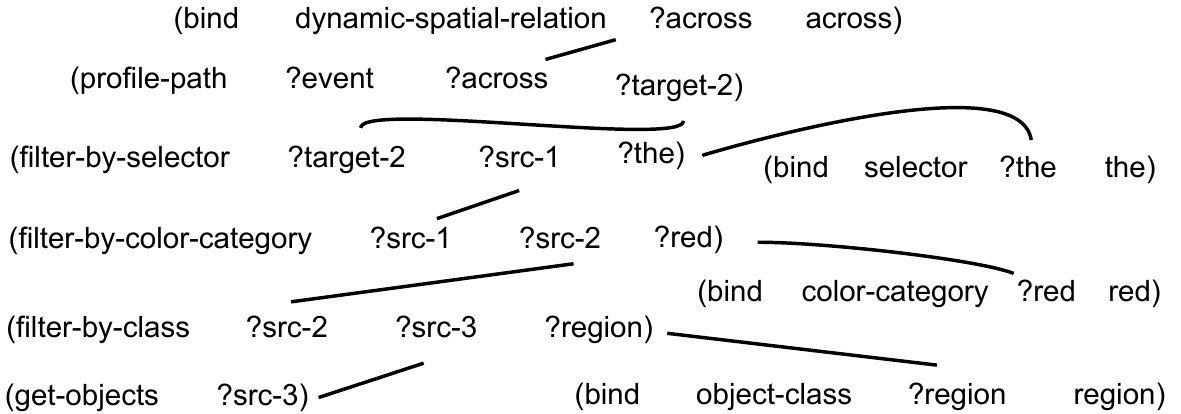}
\end{center}
\caption{Excerpt of the semantics of the 
phrase ``the block moves across the red region'' (excerpt : ``across the red region'')}
\label{f:irl-program}
\end{figure}
\subsection{Spatio-Temporal Semantics}

We model the semantics of spatial phrases using a computational cognitive semantics system called 
Incremental Recruitment Language (IRL) \cite{spranger2012irl}. 
The key idea in IRL is that semantics of natural language phrases can be 
modeled as a \emph{program} (henceforth IRL-program)  \cite{johnson1977procedural}. 
The meaning of an utterance consists of an algorithm and data pointers that when executed 
by the hearer will lead him to identify the topic (i.e. some event or object).

Figure \ref{f:irl-program} shows a graphical representation of the IRL-program 
(i.e. meaning) underlying some part of the phrase from Example \ref{e:block-moves-across}.
The IRL-program consists of 1) \emph{cognitive operations} (e.g. {\small\tt filter-by-class}) 
implementing algorithms such as categorization and 2) \emph{semantic entities} 
-- the data that cognitive operations work with. Semantic entities can be prototypes, concepts 
and categories or more generally representations of the current context, as well 
as data exchanged between cognitive operations. They can be introduced explicitly in the 
network via \emph{bind-statements}.  The statement  {\small\tt (bind dynamic-spatial-relation ?acr across)}
encodes the access to the agent-internal, dynamic spatial relation {\small\tt across} which
will be bound to the variable {\small\tt ?across}. Semantic entities are linked with
particular parameters of cognitive operations via variables (starting with {\small\tt ?}).
 In IRL-programs (meaning structures) many cognitive operations can be combined.
Most relevant for this paper are the spatio-temporal aspects of these programs.

\medskip

\noindent \textbf{Profiling} operations pick out aspects of movement events.
We implemented \emph{Source-Path-Goal} image schemas (known from Cognitive Semantics). 
The operation {\small\tt apply-path} picks out the trajectory of an event. Other operations deal 
with the source position or goal locations (e.g. {\small\tt apply-source}). In English source and 
goal are specifically marked using the prepositions ``to'' and ``from''. Profiling operations
work directly on predicates extracted by the reasoning system.

\medskip

\noindent \textbf{Dynamic Spatial Relations} are concerned with describing aspects of the path of 
an event. Here we focus on the major relations such as ``across'', ``in to'', ``out of''. 
The operation {\small\tt apply-dynamic-spatial-relations} computes whether a
event or set of events fulfills a trajectory condition, for example that the undergoer of the 
movement event moves across some region (all input parameters). This operation
checks the object relations computed by CLP(QS).

\medskip

\noindent \textbf{Static Spatial Relations} are for characterizing source and goal aspects
of movement events. We implemented operations that take care of locating
an object based on its position with respect to various landmark objects (robots and boxes),
various frames of reference (absolute, relative and intrinsic) and various spatial relations
(proximal, projective, absolute etc). The system integrates previous work \cite{spranger2012deviation}.

\medskip

\noindent \textbf{Conceptualization and Interpretation}
IRL includes mechanisms for the autonomous construction 
of IRL-programs. Agents use these facilities in two 
ways. First, when the speaker wants to talk about a 
particular scene, he constructs an IRL-program 
for reaching that goal. Secondly, a listener trying to interpret
an utterance will construct and evaluate programs, in order to find
the best possible interpretation of the utterance
(see conceptualization/interpretation in Figure \ref{f:semiotic-cycle}). 
The processes are constrained by the particular goal given to the system.
For instance, if the system needs to \emph{discriminate} an object or event - it automatically
selects features that are most dissimilar with other objects and events. 
If the goal is to \emph{describe}, then features that are most similar to the 
object or event are selected without attention to other objects.
Interpretation and conceptualization are implemented as heuristics-guided
search processes that traverse the space of possible IRL-programs
by automatic programming.

\medskip

\noindent \textbf{Mechanisms for Robustness} 
The system features a number of mechanisms for robustness.
For instance, the implementation of static spatial categories follows a 
lenient approach that increases tolerance for errors in perception \cite{spranger2012deviation}.
The most important mechanism for the purpose of this paper though is the interpretation
of partial networks. For instance, suppose the hearer only parsed a partial sentence
(because of transmission errors) and can only recover a partial IRL-program.
The system then tries to complete the network thereby generating various 
hypotheses that are tested with the perceptual data and the object relations
available at that moment in time. This completion allows hearers to understand
sentences even when there are utterance transmission errors and/or ungrammatical sentences.

\subsection{Spatial Construction Grammar}
In order to compute utterances for meaning (production) and 
meaning of utterances (parsing), we use a recent version of 
a computational construction grammar system called 
\emph{Fluid Construction Grammar} (FCG) \cite{steels2011design}.  
FCG allows to specify bidirectional mappings between meanings
and utterances in the form of a single grammar.
Robots operate a spatial grammar comprised of roughly 70 constructions 
(bidirectional rules) - primarily \emph{lexical constructions} for basic concepts 
(e.g. block, box), events (e.g. move), spatial relations (e.g. along, across, into, out of),
as well as a number of \emph{phrasal constructions}. 

\medskip

\noindent \textbf{Constructions} The most important constructions are \emph{lexical} and \emph{phrasal}.
Lexical constructions are bidirectional mappings between semantic entities 
and words. For instance, there is a lexical construction for ``across'' that 
maps {\small\tt (bind dynamic-spatial-relation ?acr across)} to the stem ``across''. 
Phrasal constructions take into account the larger syntactic and semantic context. An example is 
the adjective-noun-phrase construction, which looks for an adjective and a noun as well as a 
particular linkage of operations in the IRL-program and adds word order information. 
Similar constructions are implemented for determined noun phrases, prepositional phrases
and verb phrases.

\medskip

\noindent \textbf{Mechanisms for Robustness} The single most
important robustness mechanism for the grammar is that the system applies as many constructions as possible.
This is helpful when there are transmission errors, words can not be recognized and there
are grammatical problems with word order etc. Even in such cases, the system
will try to catch the lexical items that are recognizable in a phrase and they will be mapped to 
semantic entities, concepts etc. Moreover, fragments of utterances such as noun phrases that 
are recognizable will be processed as well. This information can be used by the semantics system 
to try and understand even phrases with errors.

\section{Evaluation and Results}

In order to evaluate the whole system we developed 
scenarios in which two robots interact with each other.
Robots interact on roughly 200 pre-recorded spatial scenes 
(similar to the one depicted in Figure \ref{f:setup}). Scenes vary
in spatial configurations of the two robots, objects, regions boxes etc.

In an interaction, one of the agents acts as the speaker, the other as 
the hearer. Roles are randomly assigned. The speaker picks some aspect
of the scene and describes it to the hearer. For instance, the speaker might 
choose to describe the path of the moving object. The speaker describes the 
scene and the hearer tries to see if this is a possible description of the scene 
from his perspective. The interaction is a success if the hearer can agree with the description.
The following details the interaction steps (see also Figure \ref{f:semiotic-cycle})

\begin{enumerate}
\item The robots perceive the scene and reason about spatio-temporal relations
of objects.
\item The speaker \emph{conceptualizes} a meaning comprised of 
dynamic or static spatial relations, and construal operations for 
describing the scene.
\item The speaker expresses the conceptualization using an English grammar. 
E.g., the speaker \emph{produces} ``the green block moves from left of you, across the red region, to right of me''.
\item The hearer \emph{parses} the phrase using his English grammar and computes
the meaning underlying the phrase.
\item When the hearer was able to parse the phrase or parts of the phrase,
he examines the observed scene to find out whether the scene satisfies the conceptualization.
\item The hearer signals to the speaker whether he agrees with the description.
\item The interaction is a success if the hearer agrees with the speaker. Otherwise
it is considered a failure. 
\end{enumerate}

There are a few important points about this setup. Most importantly, each robot
sees the world from his perspective. This means that robots
always deal with issues of \emph{perceptual deviation} \cite{spranger2012deviation}.
Robots have different viewpoints on the scene, which impacts on issues
of egocentric spatial language. For instance, ``the block to the left'' can mean
different objects depending on the viewpoint. But even on a more
basic level robots will estimate the world and its properties from their 
viewpoints. This leads to different estimations of distance and direction and 
in some cases can lead to dramatic differences in perception of the scene.
The spatio-temporal continuity of objects can be disrupted, which means that
events can be missed by some robot.

Another important aspect is that robots are not only interpreting but
also speaking using the same system. Therefore, our setup allows us to quantify 
the impact of particular algorithms on the ability of robots to communicate. 

\subsection{General Evaluation}

We evaluate the performance of the system on roughly 200 spatial scenes
on which robots interact 10000 times. Each time one of the scenes is randomly 
drawn. Each time speaker and hearer are randomly assigned some perspective.
The number of descriptions that can be generated for a scene is infinite - in particular
because agents can generate arbitrarily long descriptions. For the purpose of this paper though, we
restrict generation to simple sentences that include just 1 preposition and 2 noun phrases, e.g. 
``the object moves into the red region'' or ``the object moves from left of you''.

The simplicity constraint allows us to compute all the meanings and utterances for descriptions
of a scene from the viewpoint of any of the robots. In total we observed for this data set about 
40 different utterances exchanged  between robots. Each utterance was checked by 
3 different English speakers. All of them were syntactically correct and intelligible. 

For each interaction of two robots, we track 1) whether it was successful (SUCC), 2) how often the 
speaker was able to construe a meaning in production (CM), 3) how often the speaker produced an utterance (PU), 
4) how often the hearer parsed a meaning (PM) and 5) how often the hearer was able to interpret the meaning in 
the current scene (IM). We also do one more check, which is whether the initial meaning that the speaker 
had in mind is part of the meanings recuperated by the hearer (overlap or OL)

\begin{center}
\begin{tabular}{ |c|c|c|c|c|c|}
\hline
SUCC & CM & PU &  PM & IM & OL \\ 
\hline
.78 & .99 & .99 & .99 & .78 & .79 \\
\hline
\end{tabular}
\end{center}

Results show that in roughly 80\% of interactions, the hearer can agree
with the description. In case of failure it is more likely
to be a failure of the listener to interpret the description of the speaker (IM),
then 1) a speaker coming up with a description (CM), 2) speaker
producing an utterance (PU), or 3) the hearer failing to parse the utterance (PM).

Upon further examination we observe that in cases where communication
fails, there are perceptual problems. If we ask the hearer to 
conceptualize and produce utterances using his viewpoint on the world,
we can see that the utterance of the speaker is not actually part of those
descriptions produced by the hearer in 20\% of the cases (.79 OL). 
The reason is that hearer and speaker in some scenes extract different events. 
For instance, the hearer might miss important parts of the trajectory and
cannot agree to a description (for example ``across the red region'').

This is also confirmed by examining F-scores for utterances
and meanings. For this, we have the two robots (a and b) produce
all utterances and all meanings for a scene. We then compare utterances
and meanings. True positives are those utterances produced both by b
and by a. False negatives are utterances
produced by a AND not produced by b. False positives are 
utterances produced by b AND not by a.
\begin{center}
\begin{tabular}{ |c|c|c|c|c|}
\hline
precision & recall & f-score \\\hline
85.54 & 89.97 & 87.70 \\\hline
\end{tabular}
\end{center}

We can conclude that there are problems prior
to language processing in how the scene is perceived
and subsequently conceptualized, which leads to different
utterances being produced and then false positives and false
negative utterances subsequently. 

\subsection{Evaluation of Robustness}

Results in the previous section beg the 
question how robust the system is. In further studies, 
we manipulated the two inputs to the system: visual information 
and language. Each of these can be individually perturbed to see 
when the system breaks down.

\noindent \textbf{Visual Perturbations -- Missing Observations}
Firstly, we investigated dropping random frames of object observations
in the visual system. The camera computes object positions roughly 30 
frames per second. We randomly dropped object observations in 10\%, 
25\%, 40\%, 50\% and 75\% of frames - with 75\% meaning that on 
average 45 frames for every 60 frames are dropped. We varied
the selectivity of this effect in three cases: \emph{both} robots, 
\emph{only-a} and \emph{only-b}. For the latter conditions
only one of the robots experiences missing object perceptions.
We measured precision, recall and f-score for all utterances by a and b.

\begin{table}
\begin{center}
\begin{tabular}{ |p{2.1cm}|c|c|c|c|c|}
\hline
drop & precision & recall & f-score \\\hline
baseline & 85.54 & 89.97 & 87.70 \\\hline
\hline
10\%, both &  85.28 & 89.97 & 87.56 \\\hline
25\%, both & 85.17 & 89.87 & 87.45 \\\hline
40\%, both &  83.97 & 89.47 & 86.63 \\\hline
50\%, both & 83.99 & 89.18 & 86.50 \\\hline
75\%, both & 77.04 & 81.47 & 79.19 \\\hline
\hline
10\%, only-a &  85.35 & 89.95 & 87.59 \\\hline
25\%, only-a & 85.12 & 90.23 & 87.60 \\\hline
40\%, only-a &  84.20 & 89.96 & 86.98 \\\hline
50\%, only-a & 83.14 & 90.25 & 86.55 \\\hline
75\%, only-a & 68.19 & 91.51 & 78.15  \\\hline
\hline
10\%, only-b & 85.51 & 89.77 & 87.59 \\\hline
25\%, only-b &  85.77 & 89.63 & 87.66 \\\hline
40\%, only-b &  86.22 & 89.72 & 87.94 \\\hline
50\%, only-b & 86.21 & 87.97 & 87.09 \\\hline
75\%, only-b & 88.87 & 73.10 & 80.22  \\\hline
\end{tabular}
\end{center}
\caption{Results visual perturbations}
\label{t:results-language}
\end{table}
Results in Table \ref{t:results-language} show that the system copes 
well with missing object perceptions. 
Performance degrades gracefully 
and even when many frames of object perception
are omitted the system is still performing well.
Event with 50\% percent of frames dropped is the performance 
still comparable to baseline. Performance starts to degrade
more rapidly around 75\%. The reason for this is an interplay of various systems, but,
in particular, the resilience of the reasoning system to
missing frames of observations. The system manages to
extract stable object relations over time.

\medskip

\noindent \textbf{Visual Perturbations -- Misaligned Events}
Secondly, we investigated the impact of misalignment of events
recognized by the spatio-temporal reasoning system. For this
we measured performance of the robots in scenes where robot-a
and robot-b have different event perceptions. For instance, a 
sees the block move into a green region after it had crossed
a red region. b only sees the move into the red region, but fails
to observe the move into the green region. The dataset is a subset
of the dataset used for general evaluation. 

To see the impact of the reasoning system, we tested
two sets of agents. In one the hearer was allowed to reason about the
next possible events given his observation (wr), in the other, agents were
not allowed to reason about possible events (wor). The following table 
shows results for two populations each interacting 10000
times. 

\begin{center}
\begin{tabular}{ |c|c|c|c|c|c|c|}
\hline
& SUCC & CM & PU &  PM & IM & OL \\ 
\hline
wor & .70 & .98 & .98 & .98 & .70 & .70 \\\hline
wr & .79 & .99 & .99 & .99 & .79 & .79 \\\hline
\end{tabular}
\end{center}
Reasoning boosts success in roughly 10\% of the cases and helps establish 
agreement in description.

\medskip

\noindent \textbf{Language Perturbations}
We were also interested in impact of perturbations
of utterances computed by the speaker on the overall success.
We looked at two manipulations: word order and  missing words.

The first manipulation is to drop random words from the string
the speaker has uttered (this is similar to non-understood words).
So for instance, when the speaker said ``the block moves into the red region'',
the hearer will only see ``the moves into the red region''.
The second manipulation is to permutate words. A sentence such as
``the block moves into the red region'' might be passed to the hearer 
as ``the red moves block region the into''.

The following table shows results for 0 to 3 dropped words (d=0 to d=3)
and permutations of words (p=T - permutation; p=F - no permutation). 

\begin{center}
\begin{tabular}{ |p{1.3cm}|c|c|c|c|c|c|}
\hline
COND & SUCC & CM & PU &  PM & IM & OL \\ 
\hline
d=0, p=F & .78 & .99 & .99 & .99 & .78 & .79 \\
\hline
d=1, p=F & .89 & .99 & .99 & .99 & .89 & .79 \\
\hline
d=2, p=F & .78 & .99 & .99 & .99 & .78 & .69 \\
\hline
d=3, p=F & .82 & .99 & .99 & .99 & .82 & .70 \\
\hline
d=0, p=T & .70 & .99 & .99 & .99 & .70 & .60 \\
\hline
d=1, p=T & .74 & .99 & .99 & .99 & .74 & .60 \\
\hline
d=2, p=T & .78 & .99 & .99 & .99 & .78 & .65 \\
\hline
d=3, p=T & .83 & .99 & .99 & .99 & .83 & .70 \\
\hline
\end{tabular}
\end{center}

Results suggest that agents are well capable of dealing
with language perturbations. If anything communicative success
improves because the hearer can rearrange the words
in such a way or imagine missing words so as to make
the sentence fit his observation of the scene.

\section{Discussion}

The system presented in this paper is a fully working system 
able to interpret and produce natural language phrases with 
dynamic and static spatial relations. Such a system is useful 
for human-robot interaction about aspects of the environment.
For instance, components of these phrases can be used in question-answer
scenarios or in command-driven human-robot interfaces. Robots can
understand the need to move to a certain location. Description of regions
path, source and goal can be used to drive behavior and action
planning systems. Part of our ongoing work is to test the system 
for command language with human subjects.

This paper reviewed the proposed system primarily with 
respect to perturbations in visual processing and language
transmission. We believe that this is a fruitful way of analyzing
the robustness of Natural Language systems, something
that is often not done in the AI/Robotics community. Importantly,
we found that robustness is primarily a function of integration
of various cues from vision, reasoning, semantics and syntax.
Only if each part of the system has some notion of dealing
with perturbations can the system as a whole cope with
various robustness issues.

\bibliographystyle{named}
\bibliography{ijcai16}

\end{document}